\documentclass[english]{ccdconf}
\usepackage{amssymb}
\usepackage{amsthm}
\usepackage{amsmath}
\newtheorem{remark}{Remark}
\newtheorem{assumption}{Assumption}
\newtheorem{proposition}{Proposition}
\usepackage[linesnumbered,ruled]{algorithm2e}
\usepackage{bm}
\usepackage{cite}
\begin{document}

\title{Optimization-based Motion Planning for Autonomous Parking Considering Dynamic Obstacle: A Hierarchical Framework }
\author{Xuemin Chi\aref{amss}, Zhitao Liu\aref{amss}, Jihao Huang\aref{amss},
      Feng Hong\aref{amss}, Hongye Su\aref{amss}}

\affiliation[amss]{State Key Laboratory of Industrial, Control Technology,
Zhejiang University, Hangzhou, 110004
        \email{chixuemin@zju.edu.cn, ztliu@zju.edu.cn, jihaoh@zju.edu.cn, hogfeg@zju.edu.cn, hysu@iipc.zju.edu.cn }}
\maketitle

\begin{abstract}
This paper introduces a hierarchical framework that integrates graph search algorithms and model predictive control to facilitate efficient parking maneuvers for Autonomous Vehicles (AVs) in constrained environments. 
In the high-level planning phase, the framework incorporates scenario-based hybrid A* (SHA*), an optimized variant of traditional Hybrid A*, to generate an initial path while considering static obstacles. 
This global path serves as an initial guess for the low-level NLP problem. 
In the low-level optimizing phase, a nonlinear model predictive control (NMPC)-based framework is deployed to circumvent dynamic obstacles.
The performance of SHA* is empirically validated through 148 simulation scenarios, and the efficacy of the proposed hierarchical framework is demonstrated via a real-time parallel parking simulation.
\end{abstract}


\footnotetext{This work was partially supported by National Key R\&D Program of China (Grant NO. 2021YFB3301000); Science Fund for Creative Research Group of the National Natural Science Foundation of China (Grant NO.61621002), National Natural Science Foundation of China (NSFC:62173297), Zhejiang Key R\&D Program (Grant NO. 2021C01198,2022C01035).}

\section{INTRODUCTION}
Autonomous parking systems are a critical component of autonomous driving technologies. These systems comprise several interrelated modules such as sensing, localization, decision-making, planning, and control~\cite{paden2016survey}. This paper narrows its focus to address challenges in the planning module of autonomous parking systems.

In an autonomous parking system, the planning module is primarily responsible for generating a viable trajectory that allows the vehicle to park in a designated space without colliding with any obstacles. This trajectory is subsequently executed by a lower-level control module~\cite{motionplanningreview}. The task of trajectory planning encompasses two major concerns: comfort and collision avoidance. An optimal trajectory minimizes time, while also considering factors such as passenger comfort and vehicular stability. In this context, we propose a hierarchical framework that integrates graph search algorithms with nonlinear model predictive control methods to generate safe and efficient parking trajectories, even in constrained environments populated by dynamic obstacles.

\subsection{Graph Search-based methods}
Graph search methods are a popular choice in the realm of path planning due to their computational efficiency relative to generic optimization techniques. In this approach, the environment is discretized into a grid, within which nodes are sampled based on specific rules. Subsequently, an algorithm searches for the optimal nodes that form the desired path. Although these methods are typically computationally efficient in low-dimensional spaces, their performance deteriorates in higher dimensions. Furthermore, the resulting paths are often sub-optimal.

Conventional deterministic graph search algorithms, such as A* and Hybrid A~\cite{hybridA*}, rely on fixed motion primitives for sampling. The motion primitives in Hybrid A* are particularly well-suited to accommodate the non-holonomic constraints of vehicles. Stochastic graph search techniques like Rapidly-exploring Random Trees (RRT)\cite{RRT} and its variants\cite{RRT*,LQRRRT} employ random sampling within grids. While these methods can be more flexible, they often produce paths with curvature discontinuities, making them unsuitable for immediate use without further refinement.

\subsection{Related work}
Two primary approaches dominate the landscape of path and trajectory planning for autonomous parking: search-based methods and optimization-based methods. While the former offers computational efficiency, the latter allows for a more nuanced consideration of a vehicle's dynamic or kinematic characteristics through model-based algorithms. Optimization-based methods also have the capability to handle complex constraints such as comfort and stability through mathematical modeling.

Zhang et al.~\cite{zhangxiaojing} proposed the optimization-based collision avoidance (OBCA) algorithm, which reformulates collision avoidance as smooth constraints using duality and Slater's condition. Despite these advances, their method is not suitable for real-time applications and struggles with dynamic obstacles.

Model predictive control has been extensively employed in trajectory planning and tracking due to its ability to manage multiple constraints effectively~\cite{MPC_DQ}. Soloperto et al.\cite{robust} applied OBCA within a tube-based robust MPC framework for car overtaking simulations. However, the robustness of real-time applicability in structured roads is not well-established. Br{\"u}digam et al.\cite{SMPC} implemented a stochastic MPC scheme for overtaking maneuvers but did not address the complexities of parking scenarios that involve both forward and backward driving.

\textbf{Contributions:} This paper presents a novel hierarchical framework designed to perform real-time parking maneuvers. The contributions of our work are summarized as follows:

\begin{itemize}
\item We introduce a hierarchical framework capable of accommodating general parking scenarios in constrained environments.
\item We propose a faster variant of the traditional Hybrid A*, termed scenario-based hybrid A* (SHA*), whose computational advantages can be leveraged in various contexts.
\item Our framework explicitly accounts for dynamic obstacles, thus allowing for real-time implementation of autonomous parking maneuvers.
\end{itemize}

\section{PROBLEM DESCRIPTION}
This section presents the parking scenario under consideration and outlines the problem formulation.

\begin{figure}
  \centering
  \includegraphics[width=\hsize]{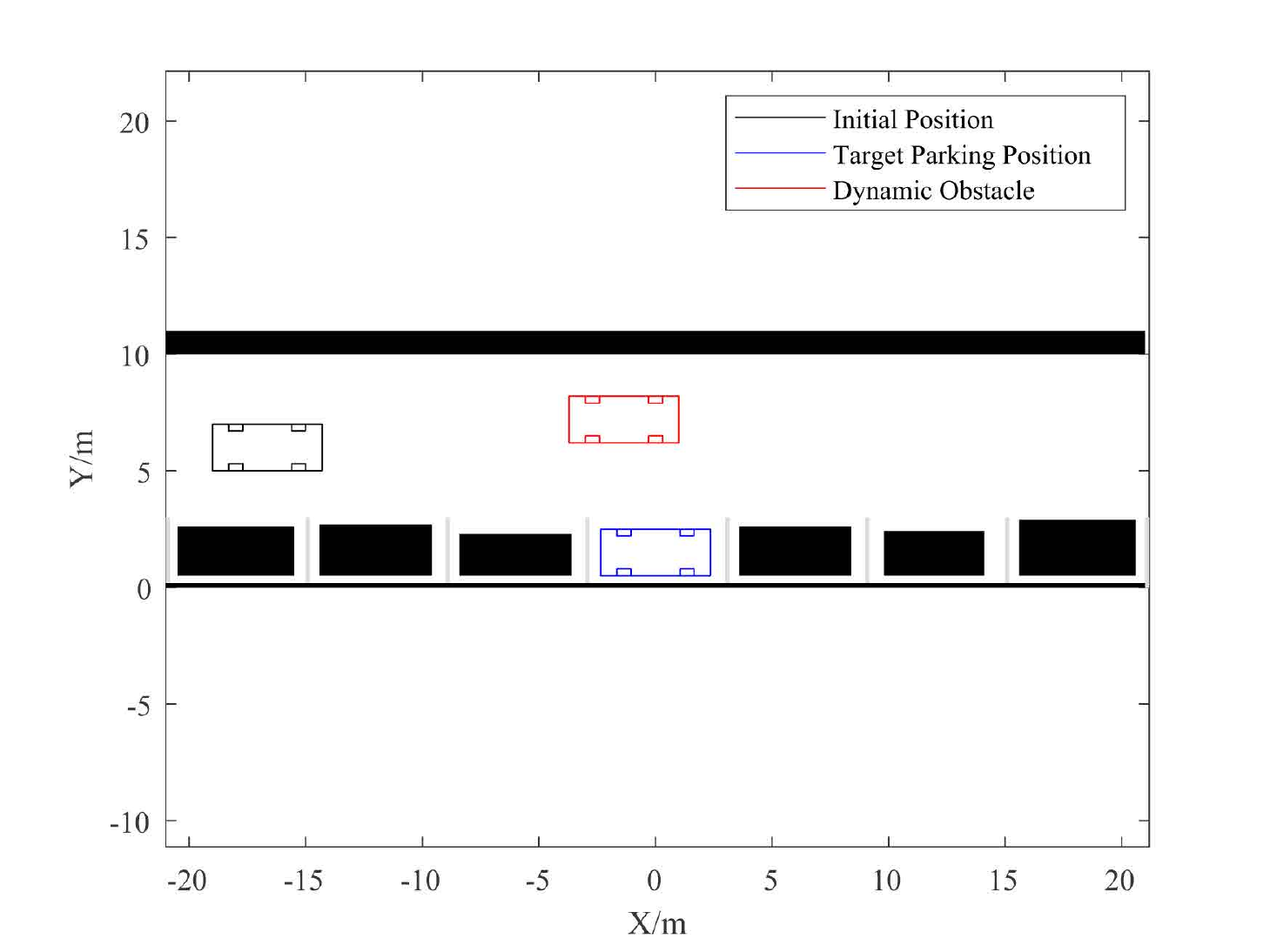}
  \caption{The AV executes a parking maneuver from start(solid black) to end(solid blue) while avoiding a dynamic obstacle(solid red)}
  \label{parking_show}
\end{figure}

We focus on a constrained parking scenario, as depicted in Fig. \ref{parking_show}.
In this environment, the AV aims to transition from a given start configuration to a predetermined goal configuration. 
A significant challenge arises from the presence of a dynamic obstacle (DO) within the parking lot. 
The AV must execute a parking maneuver that not only fulfills the objective but also avoids collision with the DO. 
Notably, the entire scenario is modeled in a 2-dimensional space.

\subsection{Vehicle Description}
\begin{figure}
  \centering
  \includegraphics[width=\hsize]{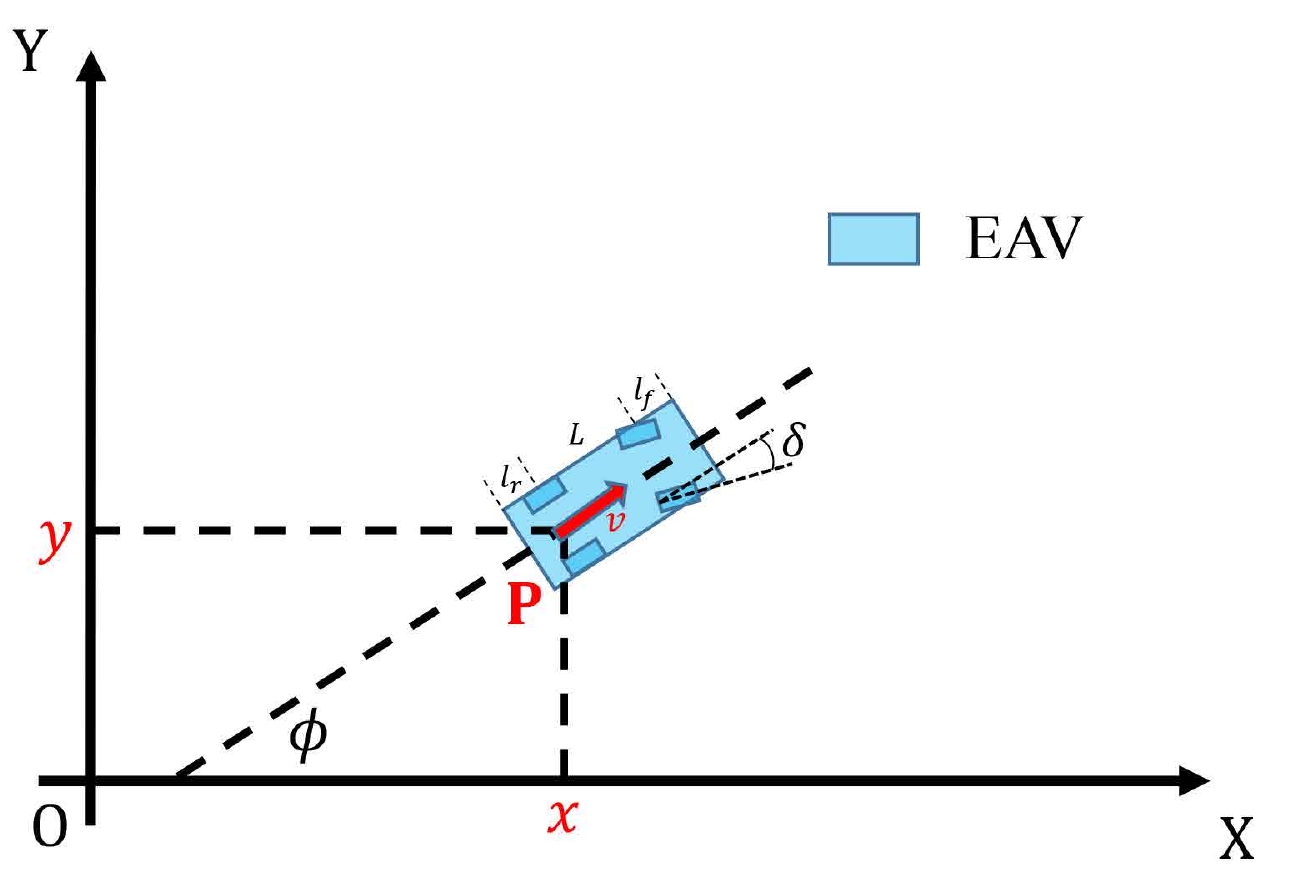}
  \caption{The proposed hierarchical parking framework}
  \label{vehicle}
\end{figure}

In the considered scenario, parking maneuvers occur at low speeds. To model the vehicle's dynamics, we employ the kinematic bicycle model, expressed as:

\begin{equation}\label{vehicle_model}
\dot{z}=\left[\begin{array}{c}
\dot{x} \\
\dot{y} \\
\dot{\phi} \\
\dot{v}
\end{array}\right]=\left[\begin{array}{c}
v \cos (\phi) \\
v \sin (\phi) \\
\frac{v}{L} \tan (\delta) \\
a
\end{array}\right], u=\left[\begin{array}{c}
\delta \\
a
\end{array}\right],
\end{equation}

where the system state $z_t \in \mathcal{Z} \subseteq \mathbb{R}^4$, $x$ and $y$ are states corresponding to the center position of the rear axle in global coordinates. State $\phi$ is the yaw angle related to the x-axis and $v$ is the velocity for the center of the rear axle. The front steering angle $\delta_f$ and acceleration $a$ are input $u_t\in \mathcal{U}\subseteq \mathbb{R}^2$. L is the length between the rear axle and front axle shown in Fig. \ref{vehicle}.

\begin{remark}
We assume that the AV parking at a low speed(i.e. less than 5m/s), therefore the tire slip angle and inertial effects can be ignored.
\end{remark}

In a tight-parking lot, a full-dimensional model allows the AV to navigate less conservative. we modeled the controlled AV as a full-dimensional object as follows

\begin{equation}
\mathbb{E}(z_t):=R(\phi_t)\mathbb{B}_0+T(x_t,y_t),
\end{equation}

where $z_t \in \mathcal{Z}\subseteq \mathbb{R}^4$ is the state of the AV at time t, $\mathbb{E}(z_t)$ is the "space" occupied by the AV, $R(\cdot)$ is rotation matrix, and $T(\cdot)$ is translation operation, $\mathbb{B}_0$ is a rectangle represented by a convex polyhedron

\begin{equation}
\mathbb{B}_0:=\left \{ y\in \mathbb{R}^2 \mid Gy\leq g \right \},
\end{equation}

where $G$ and $g$ are related to the car shape, $G = \left [ 1,0;0,1;-1,0;0,-1\right ]$ and $g = \left [L/2, W/2, L/2, W/2 \right ]$, the motion of the AV can be regarded as a polyhedron through rotation and translation.

\subsection{Obstacle Description}
In this paper, we consider $M\in \mathbb{N}, M\geq 0$ dynamic obstacles in the parking lot that the AV must avoid. DO can be described as compact polyhedrons, i.e.
\begin{equation} \label{Obs}
\mathbb{O}^{m}_t:=\left \{ y\in \mathbb{R}^2 \mid A^{m}_t y\leq b^{m}_t \right \}, m = 1,\ldots, M,
\end{equation}
where $A^{m}_t\in \mathbb{R}^{h_m\times 2}$ and $b^{m}_t \in \mathbb{R}^{h_m}$ are matrices with respect to obstacles at time $t$ and $h_m$ is the number of hyperplanes which formulates the $\operatorname{m-th}$ obstacle.

\begin{assumption}
We assumed that future $N$-steps information of DO is known. The information of future short-horizon considered predicable, which is also reasonable in reality. It can be defined as
\begin{equation}
\mathbb{O}^{m}_{k+t|t}:=\left \{ y\in \mathbb{R}^2 \mid A^{m}_{k+t|t} y\leq b^{m}_{k+t|t} \right \}, k = 0,\ldots, N,
\end{equation}
\end{assumption}

\subsection{Collision Avoidance Description}
The task of an AV is to execute the parking maneuver while avoiding all obstacles (\ref{Obs}). Formally, collision avoidance means the intersection of the space occupied by the AV and DO is empty. By using the dist function, it can be expressed as
\begin{equation} \label{avoidance}
{\rm dist}(\mathbb{E}(z_t),\mathbb{O}^m_t) \geq d_{min}, \forall m=1,\ldots,M
\end{equation}
where ${\rm dist}(\cdot)$ is the shortest distance between polyhedrons. The collision avoidance (\ref{avoidance}) is non-convex and non-differentiable in general, we exploit the proposition which provides a smooth distance and has been formulated.
\begin{proposition}
    By introducing the dual variables and the strong duality, the equivalent time-varying constraints established by \cite{zhangxiaojing} are denoted
    \begin{subequations}
    	\begin{align}
        \label{Cite.cons1} & \left(A_{k}^{m} T\left(z_{k}\right)-b_{k}^{m}\right)^{\top} \lambda_{k}^{m}-g^{\top} \mu_{k}^{m}>d_{\min }  \\
        \label{Cite.cons2} & G^{\top} \mu_{k}^{m}+R\left(z_{k}\right)^{\top} A_{k}^{m \top} \lambda_{k}^{m}=0, \\
        \label{Cite.cons3} &\left\|A_{k}^{m \top} \lambda_{k}^{m}\right\| \leq 1,  \\
        & \lambda_{k}^{m} \geq 0, \mu_{k}^{m} \geq 0, \\
        & \forall k=0, \ldots, N-1, \quad m=1, \ldots, M, \nonumber
    	\end{align}	
    \end{subequations}
    where $\lambda_{k}^{m} \in \mathbb{R}^{h_m}$ are dual variable related to $\operatorname{m-th}$ obstacles, and $ \mu_{k}^{m} \in \mathbb{R}^{4}$ are dual variables associated to the AV. (\ref{Cite.cons1})-(\ref{Cite.cons3}) explicitly reformulate (\ref{avoidance}) as constraints, and the parameter $d_{min}$ can be designed to a smaller value as the scenario gets tight.
\end{proposition}

\section{A HIERARCHICAL PARKING FRAMEWORK}
\begin{figure}
  \centering
  \includegraphics[width=\hsize]{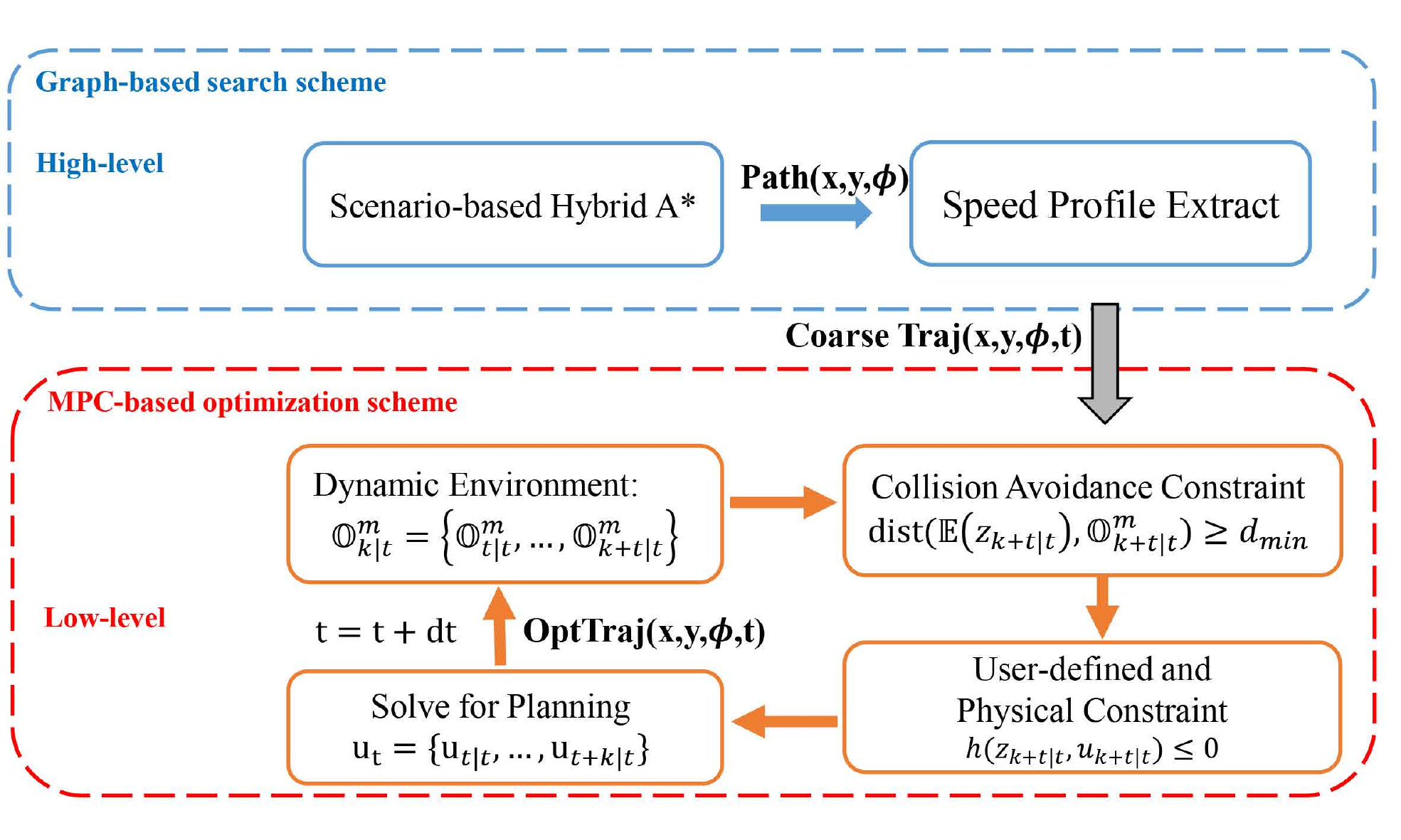}
  \caption{The proposed hierarchical parking framework}
  \label{frame}
\end{figure}

The architecture of the proposed autonomous parking system is elucidated in this section and graphically depicted in Fig. \ref{frame}. The hierarchical structure integrates two primary components: high-level path planning and low-level trajectory optimization.

In the high-level path planning phase, we utilize a scenario-based hybrid A* (SHA*) algorithm to generate an initial feasible path. The algorithm takes into account the initial and target vehicle configurations to determine a coarse path represented as $[x, y, \phi]$, devoid of temporal information.

To append velocity profiles to this coarse path, we apply a minimum-time optimal control strategy. This phase enriches the path with temporal dynamics, making it more suited for real-world applications.

The low-level controller serves as a refinement mechanism. Leveraging the minimum-time path as a reference, the nonlinear MPC algorithm incorporates DO to generate a real-time trajectory. This ensures that the resultant trajectory is not only collision-free but also maintains stability and comfort for the passengers.

By synthesizing these elements, the proposed framework achieves a comprehensive solution to the autonomous parking problem, balancing computational efficiency and real-world practicality.

\subsection{High-Level Path Planning via Scenario-Based Hybrid A* (SHA*) Algorithm}
This subsection is devoted to the high-level path planning component, focusing on the improvements introduced by employing a SHA* algorithm. The conventional HA* algorithms utilize several penalty parameters during node expansion, such as back movement, direction switch, steering angle, and steering angle gradient. Fine-tuning these penalty parameters often becomes a time-consuming endeavor. Moreover, a fixed set of parameters may not offer optimal performance across various initial parking conditions. To address these limitations, we introduce the SHA* algorithm, which adapts penalty parameters according to the specific parking scenario.

Two scenario-specific penalties have been implemented to regulate the node expansion process within SHA*. Although developed for a parking scenario, the core design strategy is transferable to other contexts as well.

\subsubsection{Yaw Angle Penalty}

\begin{figure}
\centering
\includegraphics[width=\hsize]{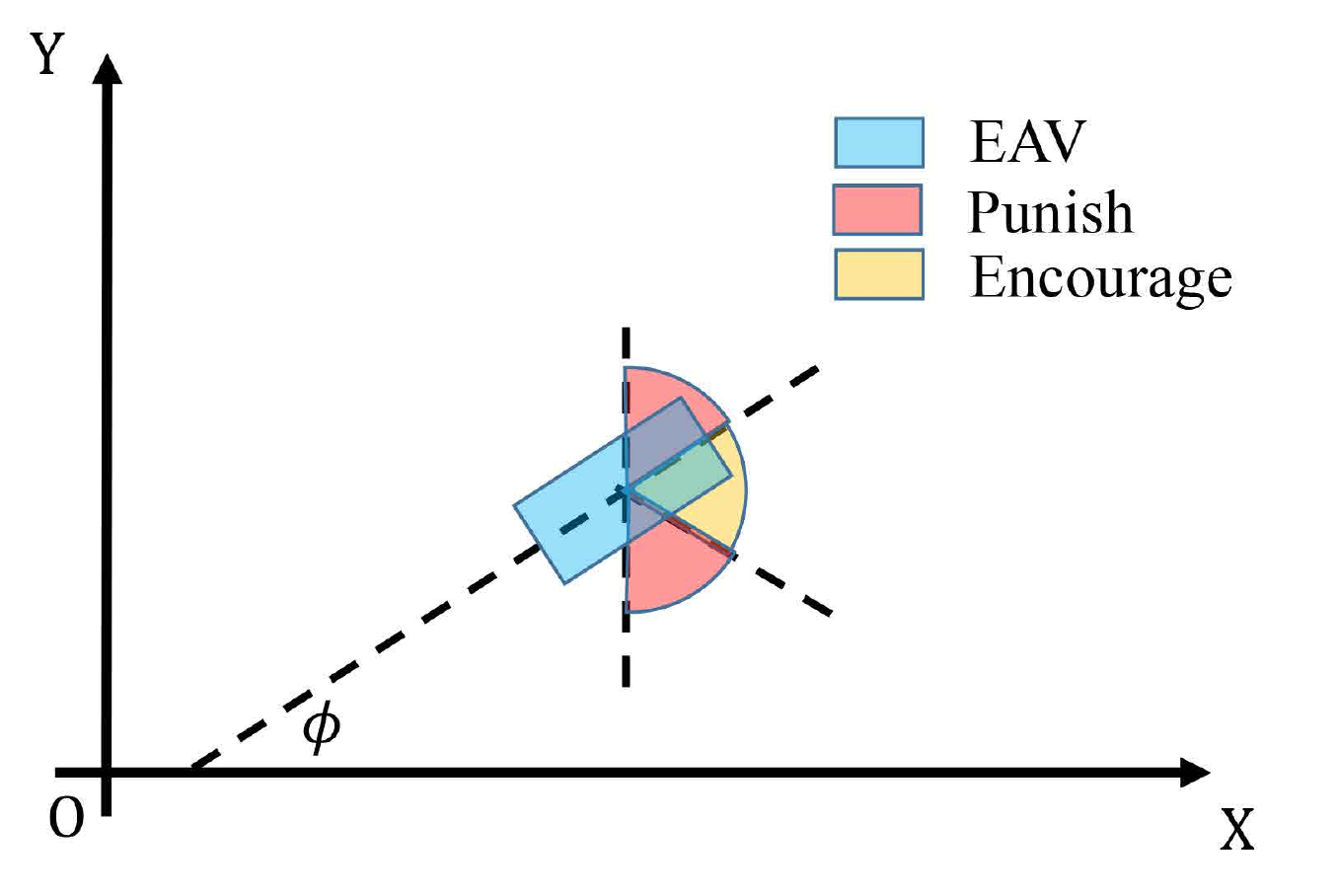}
\caption{Strategy for Minimizing Redundant Node Expansion}
\label{SHA_def}
\end{figure}

The first penalty is specified by lines 4 to 11 in Algorithm \ref{algo_SHA*}. In scenarios like parallel parking, traditional Hybrid A* algorithms tend to expand numerous nodes with yaw angles ranging between 60 to 90 degrees. Such expansions are often impractical for successful parking. To mitigate this issue, the penalty mechanism shown in Fig. \ref{SHA_def} restricts the yaw angle search to a 30-degree range. Penalties are incrementally applied for nodes with yaw angles that exceed this range.

\subsubsection{Overtravel Penalty on $x$-axis}
The second penalty aims to curtail over-extension along the 
$x$-axis during node expansion. Within the context of parking scenarios, it is generally prudent to reverse into a proximal position before finalizing the vehicle's alignment into the designated parking space. To this end, the SHA* algorithm imposes penalties on nodes based on their Euclidean distance from the intended target parking location. This mechanism biases the search algorithm towards nodes that are closer to the target location, thereby minimizing overtravel and optimizing computational efficiency.

The second penalty aims to minimize the overextension of nodes along the $x$-axis. A common approach in parking scenarios is to reverse into an approximate location before finally settling into the target position. Accordingly, we impose penalties on the nodes based on their Euclidean distance from the intended parking location. This encourages the algorithm to favor nodes that are more proximal to the target, thereby reducing overtravel.

In conclusion, the SHA* algorithm contributes to a more adaptive and efficient high-level path-planning paradigm within autonomous parking systems. By intelligently incorporating scenario-based penalties, the algorithm not only minimizes computational expenses but also provides a more focused and practical search strategy for diverse parking configurations. These refinements position SHA* as a significant advancement over traditional HA* algorithms, particularly in the realm of autonomous vehicle parking.

\IncMargin{1em}
\begin{algorithm} \SetKwData{Left}{left}\SetKwData{This}{this}\SetKwData{Up}{up}\SetKwInOut{Input}{input}\SetKwInOut{Output}{output}
	
	\Input{node $N_{current}$, penalty parameters $p$}
	\Output{next node $N_{next}$}
	\BlankLine
	
	\emph{node state information contains: $x, y, \phi , d, $}\;
    \emph{node cost information contains: $f, g, h, cost$}\;
    \emph{$count \gets 0$}\;
    \While{$ count <= 6$}{
    \eIf{$N_{parent.\phi} \geq$ limit angle  }
    {
        $N_{next.cost}\gets N_{parent.cost} + \frac{\left |\phi \right |-lim}{pi/2 - lim}*p$ ;
    }{
        $N_{next.cost}\gets N_{parent.cost}$;
    }$count \gets count + 1$
    }
    \eIf{$N_{next.x} \geq$ limit x  }
    {
        $N_{next.f}\gets N_{next.cost} + \frac{\left |x \right |-x_{lim}}{x_{target} - x_{lim}}*f_{Eu-dist}$ ;
    }{
        $N_{next.f}\gets N_{next.cost} + f_{Eu-dist}$;
    }

 	 	  \caption{The SHA* node constraint method}
 	 	  \label{algo_SHA*}
 	 \end{algorithm}
 \DecMargin{1em}

The velocity profile is acquired by an optimal control problem.
The main steps of high-level are summarized in Algorithm\ref{High-level}.

\IncMargin{1em}
\begin{algorithm} \SetKwData{Left}{left}\SetKwData{This}{this}\SetKwData{Up}{up}\SetKwInOut{Input}{input}\SetKwInOut{Output}{output}
	
	\Input{$z_{initial}$, $z_{target}$, configuration of $\mathbb{O}$ and $\mathbb{E}$}
	\Output{trajectory $z_{ref}$}
	\BlankLine
	Search global path $\left [x,y,\phi \right ]$ by SHA* \\
    Compute a coarse trajectory $\left [x,y,\phi, v\right ]$ by an OCP\\
    Coarse trajectory $z^{ref}$ is used as reference trajectory in low-level

 	 \caption{SHA*-based High-level }
 	 \label{High-level}
 	 \end{algorithm}
 \DecMargin{1em}

\subsection{Low-level: Model-based Optimization Scheme}
In the following, we present the proposed nonlinear MPC scheme. The local trajectory reference is generated w.r.t. current state of the AV. $z^{\operatorname{high-level}}_{t+k \mid t} = \left [z^{ref}_{t\mid t}, \ldots, z^{ref}_{t+k\mid t}\right ]^{\mathrm{T}}$ is generated from a start point. The start point is obtained by searching the closest point of reference trajectory to the current state.

The SHA*-NMPC formulation for parking maneuver is formulated as an NLP shown as follows
\begin{subequations}\label{MPC}
    \begin{align}
    \min _{\mathbf{z}, \mathbf{u}, \boldsymbol{\lambda}, \boldsymbol{\mu}}
                   & \sum_{k=0}^{N-1} J_1\left(z_{k+t \mid t}, u_{k+t \mid t}\right)+ J_2\left ( z_{N+t \mid t}\right)  \nonumber \\
    \label{MPC.model}   \text { s.t. } &z_{k+t+1 \mid t}=f\left(z_{k+t \mid t}, u_{k+t \mid t}\right), \scriptstyle{\forall k = 0, \ldots, N-1}  \\
    \label{MPC.set}                    &z_{k+t\mid t} \in \mathcal{Z},  u_{k+t\mid t} \in \mathcal{U}, \scriptstyle{\forall k = 0, \ldots, N}\\
    \label{MPC.x0}    & z_{t \mid t}=z_{t}\\
    \label{MPC.cons1} & {\scriptstyle \left(A_{k+t \mid t}^{m} T\left(z_{k+t \mid t}\right)-b_{k+t \mid t}^{m}\right)^{\top} \lambda_{k+t \mid t}^{m}-g^{\top} \mu_{k+t \mid t}^{m}>d_{\min }} \\
    \label{MPC.cons2} & G^{\top} \mu_{k+t \mid t}^{m}+R\left(z_{k+t \mid t}\right)^{\top} A_{k+t \mid t}^{m \top} \lambda_{k+t \mid t}^{m}=0, \\
    \label{MPC.cons3} &\left\|A_{k+t \mid k}^{m \top} \lambda_{k+t \mid t}^{m}\right\| \leq 1,  \\
    \label{MPC.h} & h\left (z_{k+t+1 \mid t}, u_{k+t \mid t} \right ) \leq 0, \\
                  & \lambda_{k+t \mid t}^{m} \geq 0, \mu_{k+t \mid t}^{m} \geq 0, \\
    & \forall k=0, \ldots, N-1, \quad m=1, \ldots, M, \nonumber
    \end{align}	
\end{subequations}%
where (\ref{MPC.model}) is vehicle model dynamics derived by discretizing (\ref{vehicle_model}) with the first-order Euler forward method over the prediction horizon, (\ref{MPC.cons1})-(\ref{MPC.cons3}) are the collision avoidance constraints. (\ref{MPC.h}) is the mechanical constraint to guarantee feasibility, stability and comfort in the physical world. For an AV and parking given by
\begin{equation}
\left[\begin{array}{c}
\Phi_{\min} \\
\mathrm{v}_{\min} \\
\delta_{\min } \\
\mathrm{a}_{\min }\\

\end{array}\right] \leq\left[\begin{array}{c}
\phi_{k+t\mid t} \\
v_{k+t\mid t} \\
\delta_{k+t\mid t} \\
a_{k+t\mid t} \\

\end{array}\right] \leq\left[\begin{array}{c}
\Phi_{\max } \\
\mathrm{v}_{\max } \\
\delta_{\max } \\
\mathrm{a}_{\max }\\

\end{array}\right], \forall k=0\ldots N
\end{equation}

In the cost function, $J_1$ is stage cost and $J_2$ is terminal cost as follows
\begin{equation}\label{parameters}
\begin{aligned}
J_1\left(z_{t+k \mid t}, u_{t+k \mid t}\right)=&\left\|z_{t+k \mid t}-z^{\operatorname{high-level}}_{t+k \mid t}\right\|_{Q}^{2}+\left\|u_{t+k \mid t}\right\|_{R}^{2} \\
&+\left\|u_{t+k+1 \mid t}-u_{t+k \mid t}\right\|_{\Delta}^{2}\\
J_2\left(z_{t+N \mid t}\right)=&\left\|z_{t+N \mid t}-z^{\operatorname{high-level}}_{t+N \mid t}\right\|_{Q_{N}}^{2}
\end{aligned}
\end{equation}

\section{SIMULATION RESULTS}
In this section, simulation results of parallel parking are provided. The simulation was conducted in MATLAB 2021 environment and performed on an MSI Titan with Intel(R) Core(TM) i7-6820HK CPU clocked at 2.7GHZ with 16GB RAM. The programming formulation language at the high level is CasADi\cite{casadi} and the low level is YALMIP. Interested readers can refer to \cite{CASADI_MORE} and \cite{lofberg2004yalmip} for more information. The NMPC of (\ref{MPC}) is solved by a general solver IPOPT. Basic simulation parameters settings are listed in Table\ref{tab1}.

\begin{table}
  \centering
  \caption{Parameters Settings}
  \label{tab1}
  \begin{tabular}{c|c|c}
    \hhline
    \textbf{Parameter}               & \textbf{Description}  & \textbf{Value} \\
    \hhline
    $\mathrm{L}$                &   length of the AV               & 4.7 m   \\ \hline
    $\mathrm{W}$                &   width of the AV                & 2 m     \\ \hline
    $\Phi$                      &   the limit of yaw rate           & 0.7 rad \\ \hline
    $\mathrm{v}$                & velocity at the center of rear    &$[-1,2]$m/s \\ \hline
    $\mathrm{\delta}$           & mechanical steer limit            & 0.6 rad \\ \hline
    $\mathrm{a}$                & acclearation bound                &$[-1,1]$m/s\\ \hline
    $\mathrm{N}$                & prediction horizon                & 5 \\
    \hhline
  \end{tabular}
\end{table}

\begin{figure}
  \centering
  \includegraphics[width=\hsize]{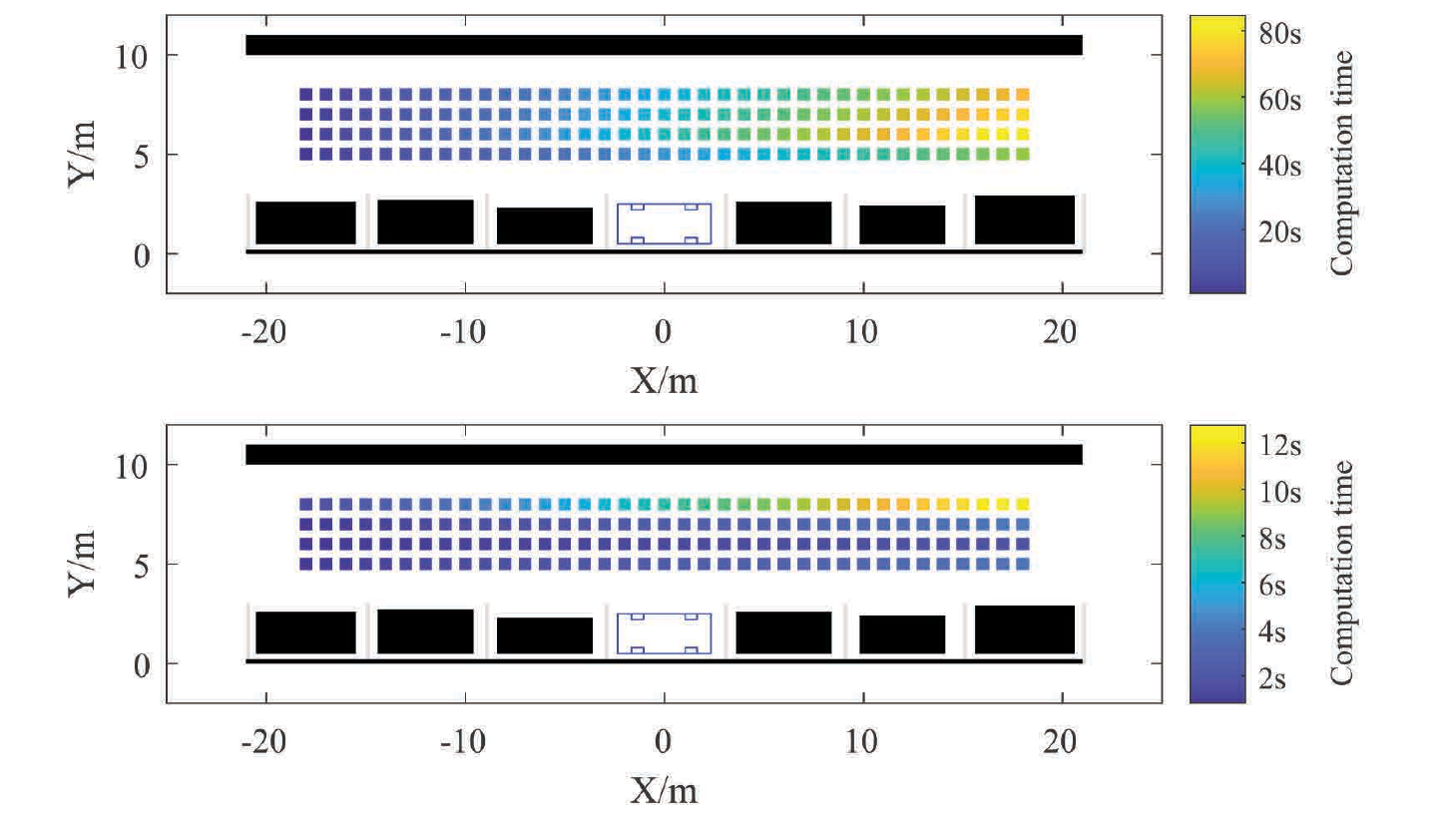}
  \caption{Computation time of different initial parking configuration for Hybrid A* and SHA*}
  \label{BOTH}
\end{figure}

To evaluate the improvement of our SHA* method, we tested 148 different initial parking position. The result compared to HA* are listed in Table \ref{tab2}. The $x$ ranges from -18 to 18 and $y$ ranges from 5 to 8 each with a increment 1. The results are shown in Fig. \ref{BOTH}.

In some cases, the computation time between HA* and SHA* is small. But in most cases, our SHA* consumed less time than traditional HA*, especially in some cases which can cause unusually long computation times for traditional HA*. It should be noted that the performance of HA* and SHA* can be improved by tuning some penalty parameters, but it is a time-wasted job and fixed parameters lack robust for all initial parking position. In contrast, our method need no parameters tuning and has a good performance as shown in Table \ref{tab2}.

\begin{table}
  \centering
  \caption{Computation Time of HA* and SHA*}
  \label{tab2}
  \begin{tabular}{c c c c}
    \hhline
                                & \textbf{min}                      & \textbf{max}              & \textbf{average}\\
    HA*                   &   0.8774 s                        & 84.6162 s                 & 9.6519 s \\ \hline
    SHA*                        &   0.8474 s                        & 12.7495 s                 & 2.1445 s \\ \hline
  \end{tabular}
\end{table}

\begin{figure}
  \centering
  \includegraphics[width=\hsize]{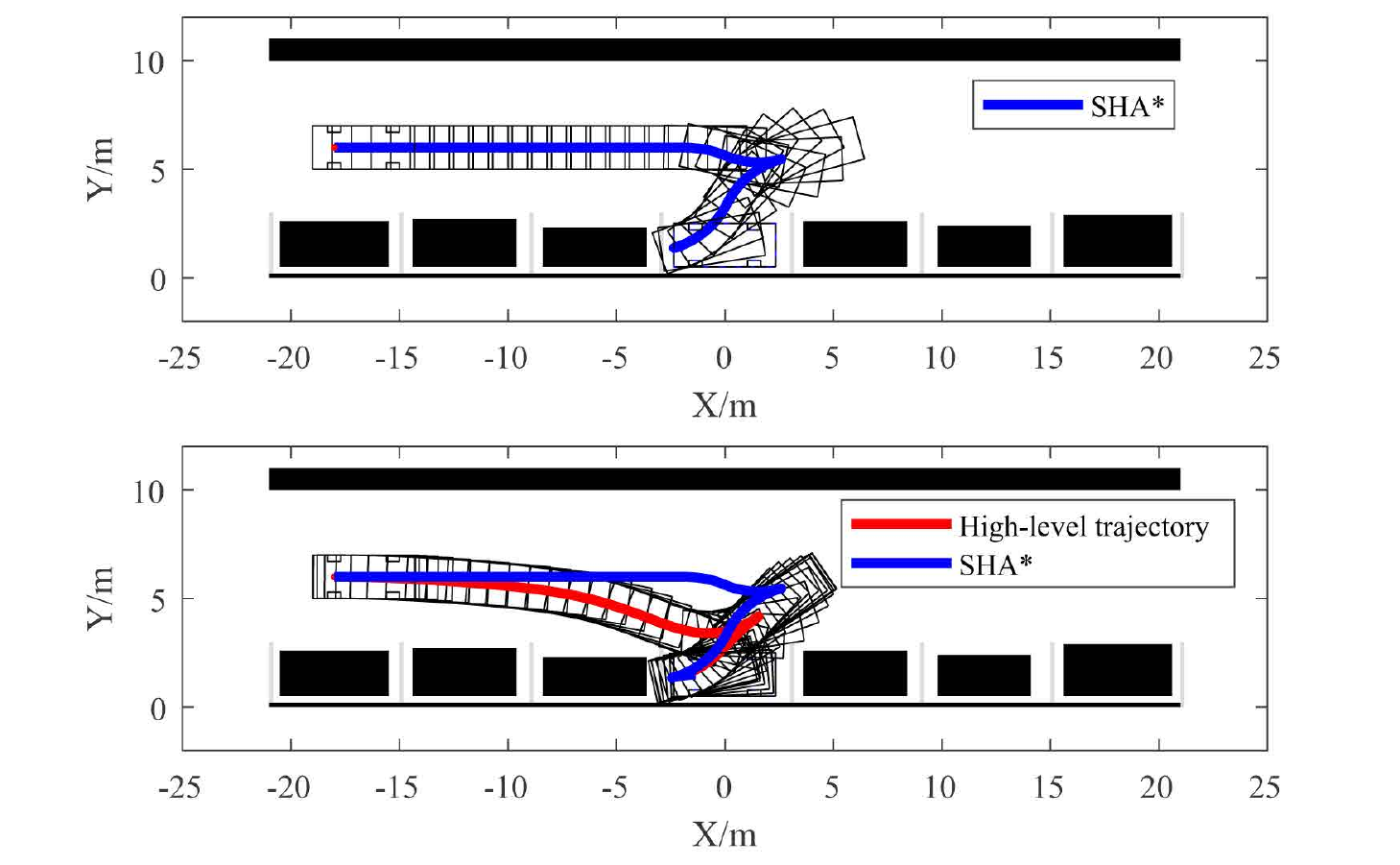}
  \caption{Trajectory planned by high-level module}
  \label{High_level_1case_fix}
\end{figure}

We take $z_{initial}=\left [ -18, 6, 0, 0 \right ]$ as a case to test our hierarchical motion planning system. Results of the high-level SHA* and time-based optimization are shown in Fig. \ref{High_level_1case_fix}. We can see that DO are ignored in high-level and the information of map and obstacles are known so the high-level can compute offline. 

\begin{figure}
  \centering
  \includegraphics[width=\hsize]{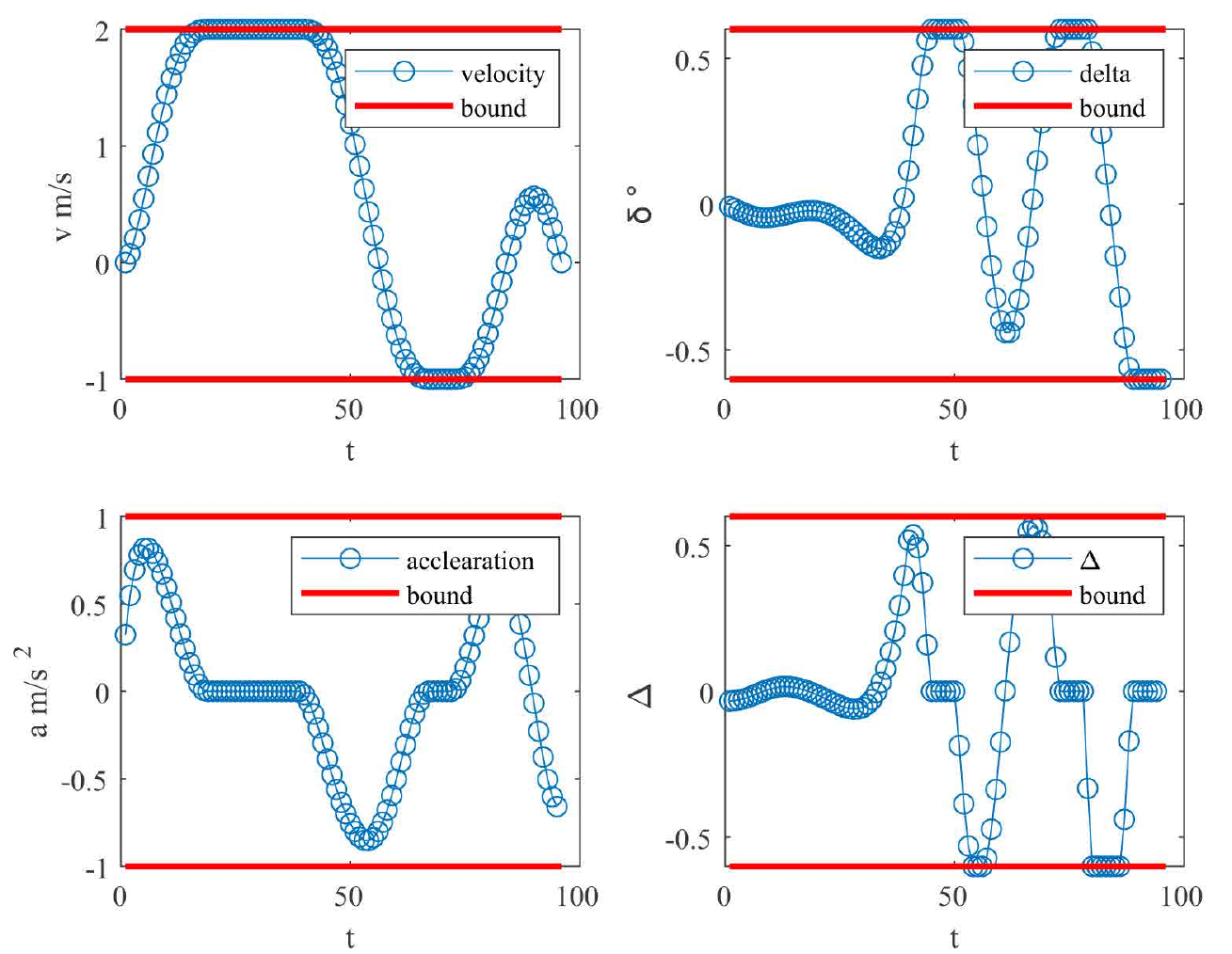}
  \caption{The important values of states from high-level}
  \label{high_level_value}
\end{figure}

Some important values of states are given in Fig. \ref{high_level_value}. The constraints in the high-level will be tighter than those in the lower module because the upper layer does not need to consider dynamic obstacles and the scenario is simpler. When considering dynamic obstacles, we tend to give more room for manipulation by using looser constraints to prevent unsolvable situations. 
\begin{figure}
  \centering
  \includegraphics[width=\hsize]{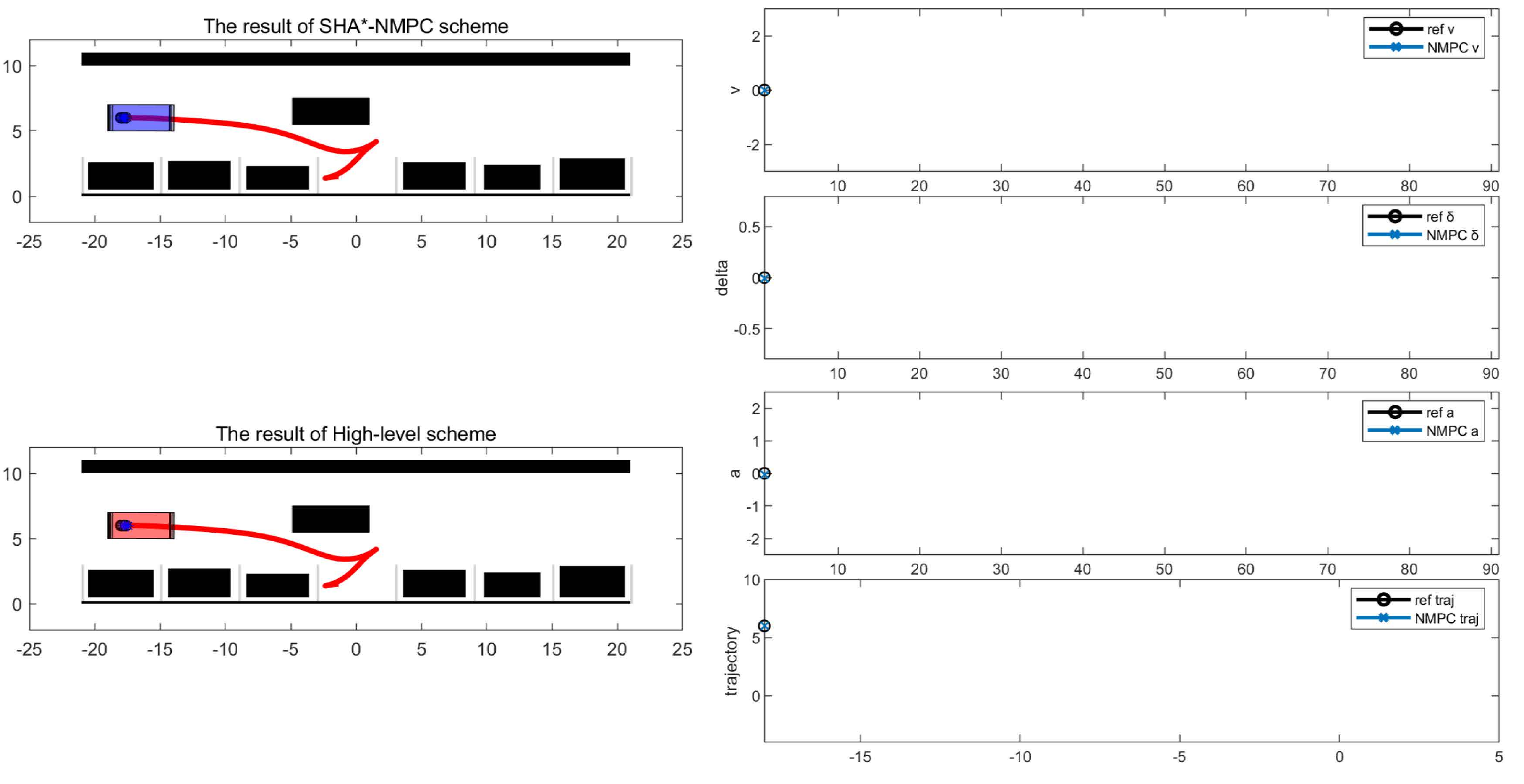}
  \caption{The configuration of autonomous parking scenario}
  \label{NMPC1}
\end{figure}

In low-level, a nonlinear MPC-based scheme is implemented to execute real-time parking maneuvers in dynamic environments. The parking scenario is shown in Fig. \ref{NMPC1}. The dynamic obstacle is moving at a constant speed of $1m/s$, from horizontal right to left. The state od the DO is $z = \left [0, 6.5, -\pi, 0 \right ]$. The DO is set to drive straight to the left in Fig. \ref{NMPC1}. The AV has to avoid the DO in the process of parking from left to right. 
\begin{figure}
  \centering
  \includegraphics[width=\hsize]{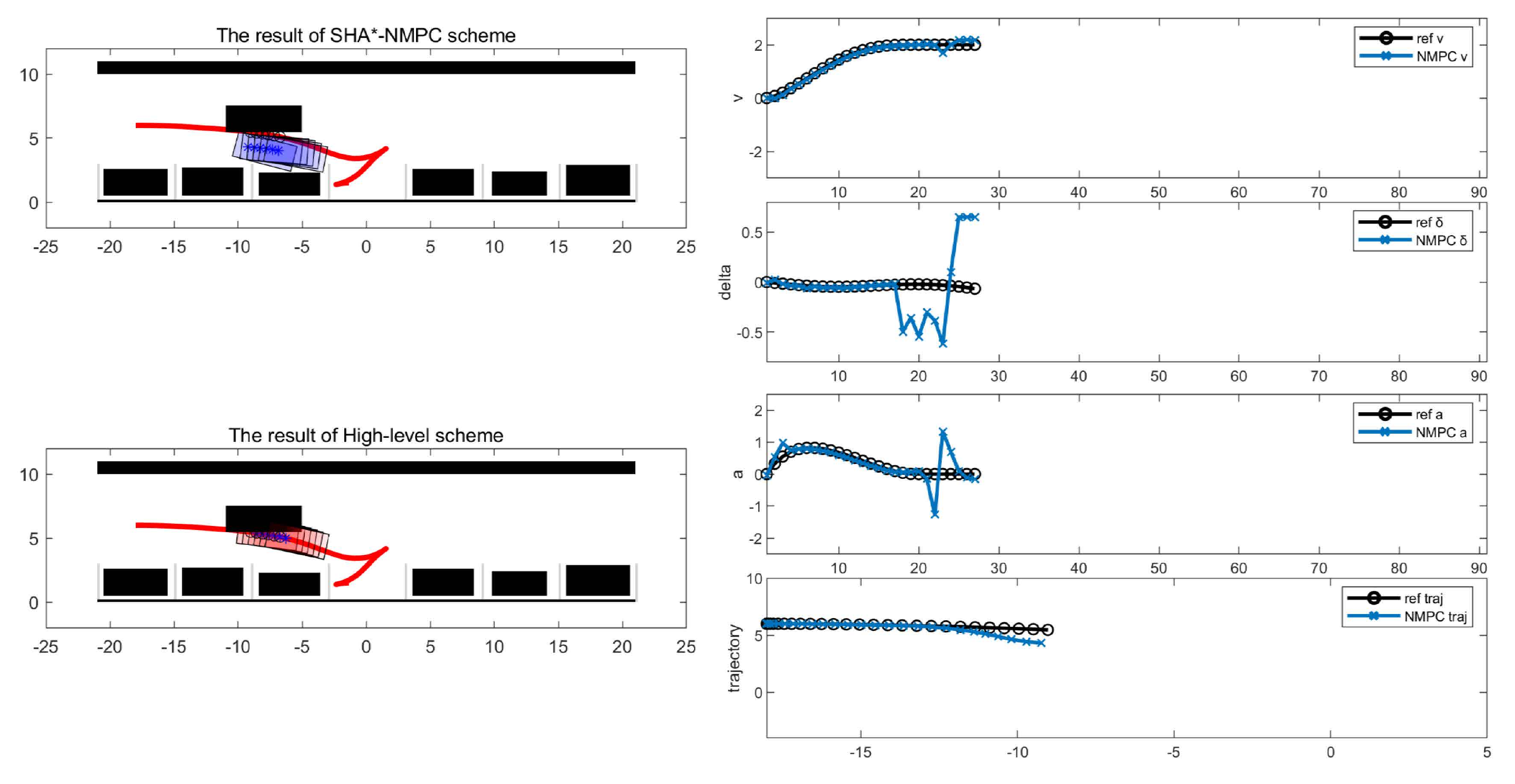}
  \caption{The generation of collision-free trajectory}
  \label{NMPC2}
\end{figure}
\begin{figure}
  \centering
  \includegraphics[width=\hsize]{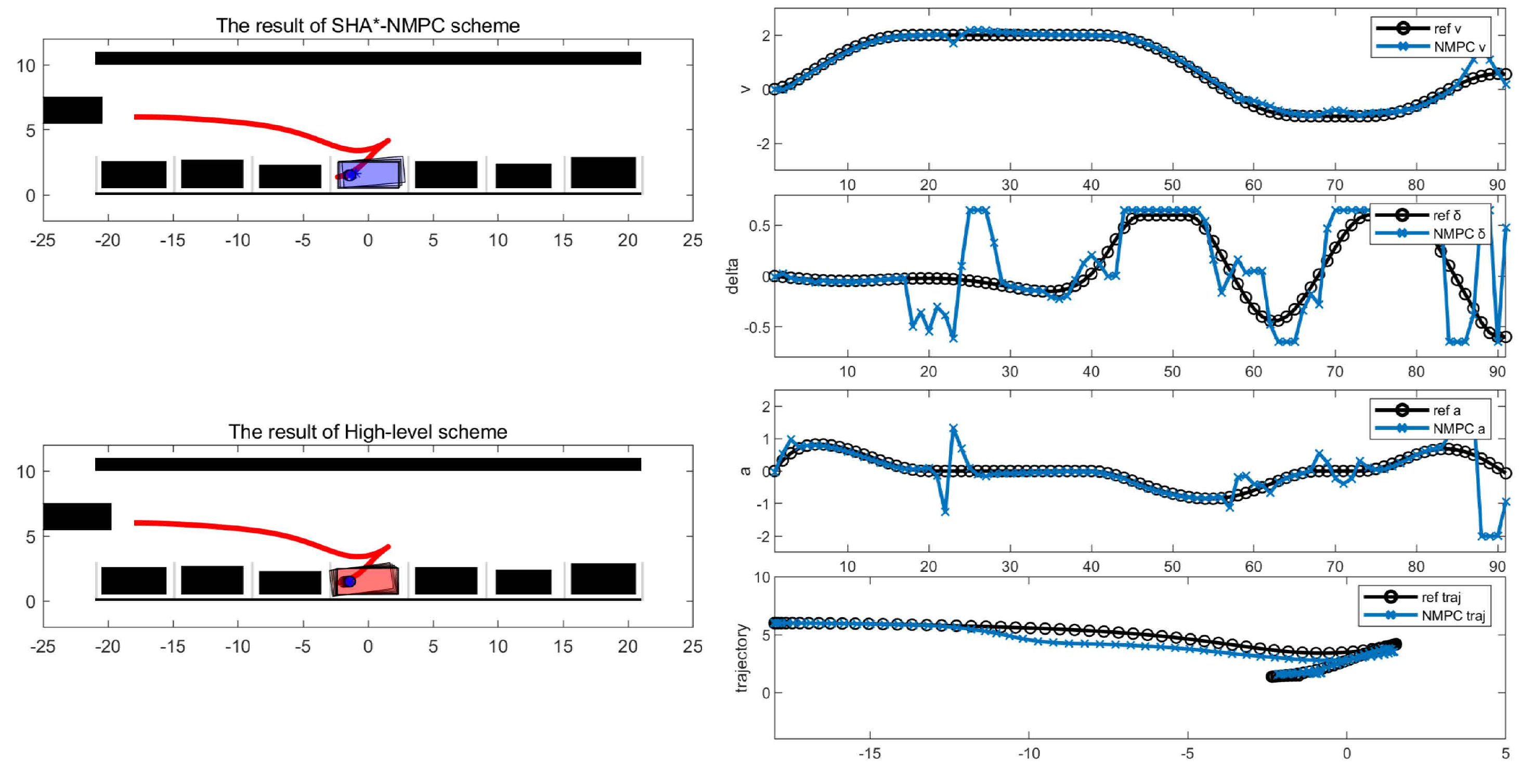}
  \caption{The final step of parking maneuver}
  \label{NMPC3}
\end{figure}
At low-level, the red reference leads to a collision shown in Fig. \ref{NMPC2}. It can be seen that, while tracking the red reference trajectory which is computed by the high-level module, in order to avoid collision with the DO, the AV was moved downward some distance. The results can also be seen in the change of the steering angle$\delta$, which deviates from the reference value in order to avoid dynamic obstacles. Contrary to this, the lower-left corner of the red AV shows the area where the collision occurred.

The final result is given in Fig. \ref{NMPC3}. From the graph, it can be learned that in the process of obstacle avoidance, the main adjustment is the steering turning angle, and because the steering angle and the amount of angle change are constrained, the acceleration variation is also needed to cope with the obstacle avoidance. After avoiding the obstacles, AV quickly followed the reference trajectory and finally achieved parking. 

The choice of parameters in (\ref{parameters}) can influence the behaviors in collision avoidance. Specific parameters are given in Table \ref{NMPC_para_tab}. In the parallel parking behavior, we considered more the angle and horizontal coordinates to follow, and the vertical coordinates and speed are of secondary importance. If the autonomous parking system is planned with the wrong angle, it is likely not a qualified parking behavior. It should be noted that the uncertainty and stochastic errors of dynamic obstacles are not considered in this paper, and these will be considered in future work.
\begin{table}
  \centering
  \caption{NMPC Weight Matrix Parameters Settings}
  \label{NMPC_para_tab}
  \begin{tabular}{c|c|c}
    \hhline
    \textbf{Parameter}               & \textbf{Description}  & \textbf{Value} \\
    \hhline
    $\mathrm{Q}$                &   stage weight of 4 states               & $diag [4,1,2,5]$   \\ \hline
    $\mathrm{R}$                &   weight of 2 control states  
    & $diag [0.01,0.01]$     \\ \hline
    $\mathrm{\Delta}$                & weight of degree of inputs    
    &$diag [0.1,0.1]$              \\ \hline
    $\mathrm{Q}_N$           & terminal weight           
    & $diag [8,2,4,10]$ \\ \hline
    \hhline
  \end{tabular}
\end{table}

\section{CONCLUSION}
This paper has introduced SHA*-NMPC, a novel hierarchical framework designed for the autonomous parking maneuvering of AVs. The efficacy and robustness of the SHA* algorithm within the framework have been validated through 148 diverse initial parking configurations. Moreover, the practicality of the proposed framework has been demonstrated via a numerical simulation for parallel parking. Comprehensive results and analyses substantiate the strong performance of the implemented system.

However, there are limitations to our approach. The primary drawback lies in the computational complexity of the nonlinear optimization problem formulated in the low-level control scheme. While we employ warm starting techniques to accelerate the optimizer's convergence, challenges arise when incorporating more intricate Dynamic Obstacles (DOs). For example, when faced with high-speed DOs, our system may necessitate advanced strategies such as early yielding or circumvention, which could require extending the prediction horizon and consequently increasing computational time. Furthermore, although duality theory can effectively handle non-smooth problems, it introduces additional variables, complicating the optimization process.

Future work will aim to enhance the safety features of our system. One avenue of exploration is the development of intelligent algorithms capable of generating constraints earlier than the current prediction horizon. Additionally, the adoption of probabilistic optimal control strategies may provide a robust framework for handling uncertainties. Linearization techniques also hold promise for significantly reducing computational burden by reformulating the nonlinear optimization problem as a quadratic optimization problem.

\bibliographystyle{IEEEtran}
\bibliography{main}
\end{document}